\documentclass[journal]{IEEEtran}

\usepackage{cite}
\usepackage[pdftex]{graphicx}

\usepackage{array}
\usepackage[caption=false,font=footnotesize]{subfig}
\usepackage{stfloats} 
\usepackage{url}
\usepackage[english]{babel}
\usepackage{bm,amssymb,amsmath,color,algorithm,algorithmic,url}
\usepackage{flushend}
\usepackage{hyperref}
\usepackage{verbatim}

\begin{document}


\title{Pose-Based Tactile Servoing:\\ Controlled Soft Touch using Deep Learning}
\author{Nathan~F.~Lepora* and~John~Lloyd*
\thanks{N. Lepora and J. Lloyd are with the Department
of Engineering Mathematics, Faculty of Engineering, University of Bristol and Bristol Robotics Laboratory, Bristol, UK. e-mail	{\tt\footnotesize \{n.lepora, jl15313\}@bristol.ac.uk}.}
\thanks{* NL and JL contributed equally to this work.}
}

\maketitle

\begin{abstract}
This article describes a new way of controlling robots using soft tactile sensors: pose-based tactile servo (PBTS) control. The basic idea is to embed a tactile perception model for estimating the sensor pose within a servo control loop that is applied to local object features such as edges and surfaces. PBTS control is implemented with a soft curved optical tactile sensor (the BRL TacTip) using a convolutional neural network trained to be insensitive to shear. In consequence, robust and accurate controlled motion over various complex 3D objects is attained. First, we review tactile servoing and its relation to visual servoing, before formalising PBTS control. Then, we assess tactile servoing over a range of regular and irregular objects. Finally, we reflect on the relation to visual servo control and discuss how controlled soft touch gives a route towards human-like dexterity in robots.
\end{abstract}

\IEEEpeerreviewmaketitle

\section{Introduction}

\IEEEPARstart{T}{he} human tactile sense enables us to use our hands to manipulate and explore our surroundings. The extent of our handiwork is a uniquely human capability that has, for better or worse, transformed the world around us: all of our technology stems ultimately from devices made by hand. An artificial tactile sense and the capability to explore and manipulate the surroundings will enable robots to perform manual tasks currently needing human labour. This has been the goal of tactile robotics research for half a century~\cite{harmon_automated_1982}. 

However, simple tactile tasks that are routinely performed by humans still elude robots, such as feeling around everyday objects. While there have been major advances in tactile sensors and their integration into robot hands, their capabilities fall far short of human dexterity. In our view, a necessary yet underdeveloped ability is to control a soft tactile sensor to maintain contact while sliding over a surface. This is the robot analogue of tracing our fingertips over objects, called tactile servo control~\cite{berger_using_1991,chen_edge_1995,zhang_control_2000,li_control_2013,lepora_exploratory_2017,kappassov_touch_2020}. Without the ability to control how a soft fingertip interacts with an object, it is difficult to imagine how more complex tasks involving touch could be achieved.

Although there have been intermittent advances in tactile servo control, both the development of the field and the volume of research lag far behind visual servo control. The modern view of visual servoing emerged in the 1990s with two archetypal schemes: image-based and pose-based visual servo control~\cite{chaumette_visual_2006}. However, almost all developments in tactile servo control have been based on image-based servoing~\cite{yoshikawa_tactile_1994} using image features from hard, planar tactile arrays~\cite{chen_edge_1995,zhang_control_2000,li_control_2013,kappassov_touch_2020}. In this paper, we describe a framework based on the second archetypal scheme, pose-based tactile servo control, and apply it to a soft tactile sensor.

\textcolor{black}{Why is pose-based servo control suited to tactile sensing? The application of deep learning to tactile sensing enables accurate pose estimation of object features such as edges or surfaces in a manner that can be robust to the complex soft and frictional interactions from contacting the object~\cite{lepora_pixels_2019,lepora_optimal_2020}. Since pose-based servo control is built on accurate pose estimation, it therefore constitutes a natural method for controlled soft touch using deep learning.}

Overall, the main contributions of this work are to:
\begin{enumerate}
	\item Formalise {\em pose-based tactile servo} (PBTS) control, in which a model that relates tactile images to sensor poses is used to control the sensor pose directly. 
	\item Implement PBTS control with a soft tactile sensor and a model trained with deep learning to be insensitive to the sensor shear that occurs during soft interactions~\cite{lepora_optimal_2020}.
	\item Test PBTS control by sliding over 3D surfaces and edges. Regular objects were used to measure performance and irregular objects to demonstrate generality.
\end{enumerate}
All experiments were done with the BRL TacTip, a soft biomimetic optical tactile sensor~\cite{ward-cherrier_tactip_2018,lepora2021soft}. although we expect the approach will apply to other soft tactile sensors.


\section{Tactile servo control}\label{sec:2a}

In control engineering, a {\em servomechanism} (shortened to `servo') is an automatic device that uses error-sensing negative feedback to correct the action of a mechanism. The field of robotics relies on using servomotors (shortened to `servos') that precisely control their motion and final position. 

\begin{figure*}[t!]
	\centering
	\begin{tabular}[b]{@{}c@{}c@{}}
		\textbf{(a) Image-based servo control} & \textbf{(b) Pose-based servo control}\\
		\includegraphics[width=\columnwidth,trim={165 250 315 150},clip]{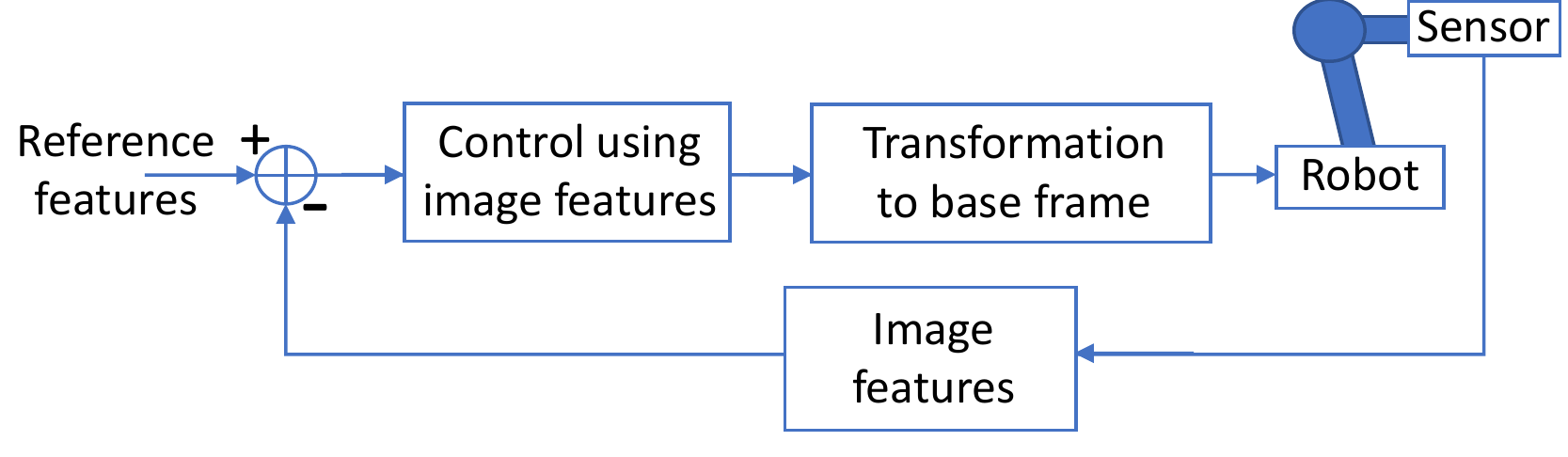} & \includegraphics[width=\columnwidth,trim={165 255 315 150},clip]{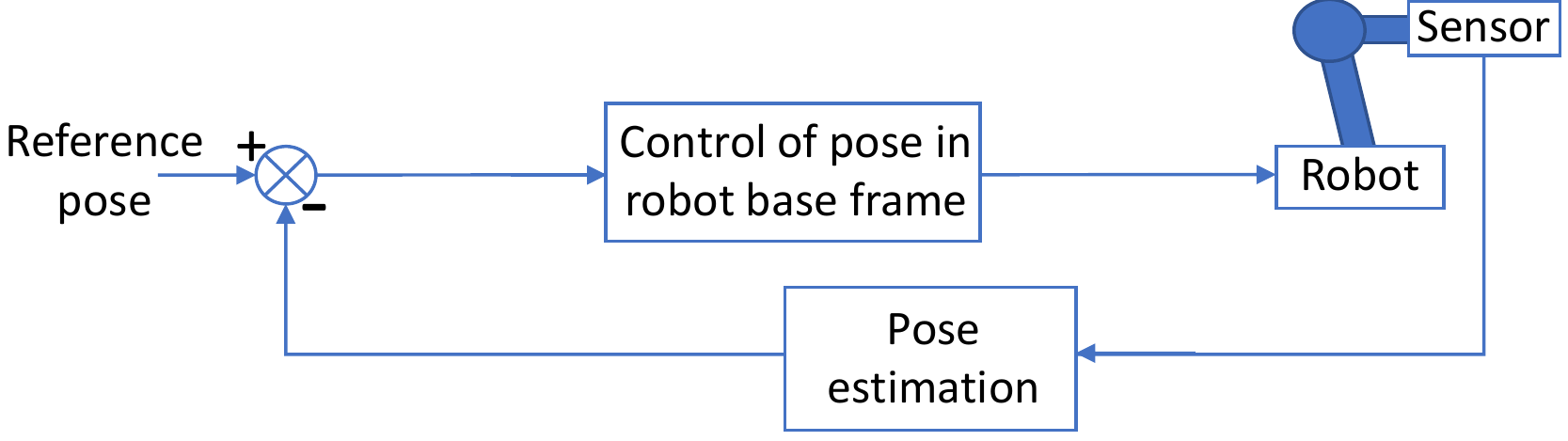}\\
	\end{tabular}
	\caption{Two archetypal schemes for feedback control of sensorized robots: (a) image-based and (b) pose-based servo control (based on~\cite[Figs 5,6]{hutchinson_tutorial_1996}). The schemes differ in their use of the sensor data: to generate image features or estimate pose, and thus whether the control is in the sensor frame of the image features or in the base frame of the robot (also called Cartesian space).}
	\label{fig3}
	\vspace{-.5em}
\end{figure*}

In modern robotics, {\em visual servo control} has received a huge amount of interest, and is based on using computer vision data in the servo loop to control the motion of a robot~\cite{chaumette_visual_2006}. Research on visual servoing dates back four decades~\cite{hill_real_1979} with the modern view emerging in the 1990s~\cite{espiau_new_1992,wilson_relative_1996,hutchinson_tutorial_1996}. A fundamental aspect of the formalism is that it distinguishes two archetypal schemes, known as {\em image-based visual servo} (IBVS) control and {\em pose-based visual servo} (PBVS) control, which differ in the type of feedback error used in the controller. The image-based scheme controls a feature vector extracted from the 2D image in the sensor frame of that image, whereas the pose-based scheme controls the camera pose directly in the base frame (Cartesian space) of the robot (\autoref{fig3}). 

This distinction between image-based and pose-based control has helped the research field of robot vision progress, with pros and cons of the two schemes (and hybrid schemes) still being investigated. The image-based scheme benefits from using image features that can be extracted straightforwardly in the control loop ({\em e.g.} image moments or key points on the image). However, the controller outputs a change in feature vector, which must be transformed to a motion of the camera, and this inverse transformation can be challenging to compute, usually with an approximate `interaction matrix' for a given scenario. The pose-based scheme benefits from controlling the camera position and orientation directly, but requires a 3D model of the object to estimate its pose from the camera image, which likewise can be difficult to calculate.
	
{\em Tactile servo control} has received far less attention and its fundamentals are still being developed. 
The first complete treatment on control of contact via tactile sensing was developed two decades ago by Zhang and Chen~\cite{zhang_control_2000}, building upon earlier work on tactile edge tracking~\cite{berger_using_1991,chen_edge_1995} and object manipulation~\cite{yoshikawa_tactile_1994,son_tactile_1996}. The control of contact relied on extracting tactile features as the input to a controller whose output fed through an `inverse tactile Jacobian' that relates these features in the image frame to the sensor motion in the base frame (Cartesian space) of the robot. Extraction of the tactile features relied on a finite element model of the sensor to infer the normal force and centre of contact from a tactile image. Evidently, this scheme is the tactile analogue of image-based visual servoing~\cite{yoshikawa_tactile_1994,kappassov_touch_2020}. Hence, we refer to it as {\em image-based tactile servo} (IBTS) control, where we interpret a {\em tactile image} as data from an array of tactile elements called taxels or tactels, which are akin to pixels of a visual image.
	
Tactile servo control had a hiatus for a decade before development continued on a control framework for tactile servoing~\cite{li_control_2013}. Li {\em et al} simplified Zhang and Chen's treatment by using a hard planar tactile array from which relevant tactile features can be estimated and controlled directly: a 2D centre of contact, normal force and orientation (from the centroid, area and moments of the tactile image). The inverse Jacobian that relates the controller output to the sensor motion was approximated with a linear transformation, analogous to the interaction matrix in IBVS control. By setting this matrix, they demonstrated tactile servoing tasks from normal force control and tracking a horizontal contact to rolling around a cylinder and sliding along a horizontal cable~\cite{li_control_2013}. Recently, Kappassov {\em et al} extended this control scheme to six features, by also including the centre of pressure, to control the sensor pose on more complex 3D tasks~\cite{kappassov_touch_2020}. With an appropriate inverse Jacobian matrix, the range of tasks was extended to slide along a bent bar in 3D and to orient to balance a tray.
	
Around the same time, another scheme was developed for tactile contour following around planar objects, using a curved, soft tactile sensor (an iCub fingertip) with a Bayesian predictor of 2D sensor pose~\cite{martinez-hernandez_active_2013,martinez-hernandez_active_2017}. Although the sensor motion was not described in standard control theory terminology, it may be considered as proportional position control applied to the sensor pose under discrete tapping contacts. This control terminology was used in a more recent application of this scheme to the BRL TacTip~\cite{lepora_exploratory_2017}. More recently, this servo control was improved by replacing the original Bayesian {predictor} with a convolutional neural network to achieve controlled 2D sliding motion around various complex objects~\cite{lepora_pixels_2019}. A key advance was that the neural network was tuned to be insensitive to motion-induced shear of the sensor surface, which is necessary for robust pose estimation with a soft tactile sensor such as the BRL TacTip.

\begin{figure*}[t!]
	\centering
	\begin{tabular}[b]{@{}c@{}}
		\textbf{Pose-based tactile servo control}                                             \\
		\includegraphics[width=2\columnwidth,trim={60 245 60 165},clip]{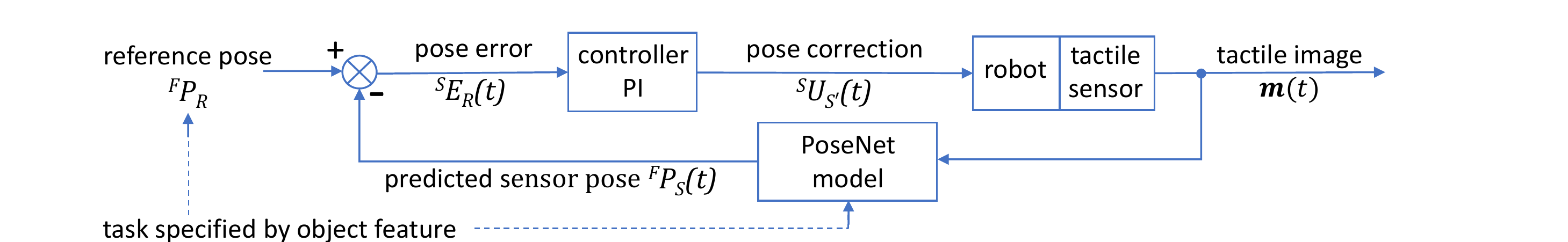}
	\end{tabular}
	\caption{Pose-based tactile servo control loop. The pose error ${}^SE_R\in{\rm SE}(3)$ between the sensor pose and a reference pose feeds into a PI controller, which drives a pose correction ${}^SU_{S'}$ of the sensor. The sensor then acquires a new tactile image $\bm m$, which the PoseNet model uses to predict the sensor pose that feeds back into the pose error. The reference pose and predicted sensor pose are defined with respect to the contact frame for an object feature (an edge or surface). The control loop iterates so that the sensor explores the object by servoing according to the predicted sensor pose relative to the object feature.}
	\label{fig1}
	\vspace{-.5em}
\end{figure*}

The present work follows from these latter studies, which we now consider as {\em pose-based tactile servo} control. Analogously to pose-based visual servoing, a model that relates tactile images to sensor poses is used to control the sensor pose directly (\autoref{fig1}). Training this model using deep learning enables discovery of latent tactile features that are insensitive to shear, which is necessary for robust and accurate servo control using soft tactile sensors.

\section{Tactile pose estimation}

In considering contact-based pose estimation, one can distinguish between tactile sensing and force/torque sensing of local object pose. A passive fingerpad attached to a force/torque sensor can detect surface pose by the method of `intrinsic contact sensing'~\cite{bicchi_contact_1993}, which models how forces determine the contact location on the fingertip surface and the local normal to the surface under contact. This method has been improved to encompass deformable fingertips, enabling accurate tracing over curved objects via a combination of normal force control and guided transverse motion~\cite{liu_finger_2015}; moreover, the object shape can be explored with methods that implicitly model the surface~\cite{rosales_gpatlasrrt_2018,driess_active_2019}.  However, intrinsic contact sensing has not been applied to other object features such as edges, which would require the estimation of edge pose from a relatively low-dimensional force/torque signal.

Tactile arrays provide high-dimensional information about surface shape. However, there are challenges in mapping the tactile image to the local pose of an object feature: (i) for soft curved tactile sensors like our fingertips, the map from the tactile image to object feature pose will be complex and highly nonlinear; (ii) more fundamentally, for soft interactions, the tactile image depends not only on the pose of the object feature, but also the history of how that object feature was contacted, for example in the shear of the soft sensor surface as it moved into its pose~\cite{aquilina_shear-invariant_2019,lepora_pixels_2019}. In our view, these difficulties have confined the use of robot touch to very primitive tasks compared with the fine motor capabilities of humans~\cite{lepora_optimal_2020}.

Past work on tactile servoing by Li {\em et al} and Kappassov {\em et al} avoided these problems by using a hard, planar tactile array~\cite{li_control_2013,kappassov_touch_2020}. In this special case, it is possible to extract tactile features from the tactile image ({\em e.g.} centre of contact and image moments) that linearly relate to the pose of the object feature and are independent of the contact motion. However, if the tactile sensor is compliant or non-planar, as is typical of tactile sensors like our fingertips, then these simple relations do not hold. Moreover, for all tactile sensors, including planar arrays, the shape and curvature of the object features will affect the tactile image, and hence its relation to the pose.

In this paper, we apply recent progress in using deep learning with the BRL TacTip to estimate the pose of object features with a soft curved tactile sensor~\cite{lepora_optimal_2020}. Specifically, deep learning can be used to train accurate `PoseNet' models of object surface and edge features, such that those models are insensitive to nuisance variables including the motion-dependent shear due to soft interactions. Our method involves using representative sensor motions as unlabelled perturbations of the training data. For more background on deep learning for tactile pose estimation, we refer to \cite{lepora_optimal_2020}.


\section{Pose-Based Tactile Servo Control}\label{sec:3}

\subsubsection{Pose error}
\textcolor{black}{The primary concept underlying pose-based servoing is the error between the sensor pose and a reference, which will be minimized by the controller. Following a classic review of visual servo control~\cite{chaumette_visual_2006}, we define this pose~error~as} 
\begin{align}
\label{eq:1}
{}^SE_R(t) = {}^SP_F(t)\,{}^FP_R = {}^FP^{-1}_S(t)\,{}^FP_R,
\end{align}
which is specified as a transformation to move the observed sensor pose ${}^FP_S(t)$ to a reference pose ${}^FP_R$ in the same coordinate frame $F$ at time $t$. These poses and pose errors are elements $T\in{\rm SE}(3)$ of the special Euclidean group of translations and rotations, which can be represented as $T=(x,y,z;\alpha,\beta,\gamma)$ in Cartesian coordinates $(x,y,z)$ and Euler angles $(\alpha,\beta,\gamma)$, in a common notation with~\cite{lloyd2021goal}. Overall, the pose error ${}^SE_R(t)$ specifies the reference sensor pose in the current sensor frame, whose components can be paramaterized in Cartesian coordinates and Euler angles.

This treatment of pose as a transformation is important because in this context it is incorrect to calculate the error by subtracting two vectors. For 3D servo control, the subtraction-based error may be a good approximation at small angles, but successive operations can build up large undesired motions. 

\subsubsection{Edge and surface features}\label{sec:3a2}
In practice, we consider the sensor pose ${}^FP_S(t)$ and reference pose ${}^FP_R$ in the coordinates of a local \textcolor{black}{contact} frame $F$ of an object feature. These object features are local regions of an object that have a specific geometry such as an edge or surface. (Note that these object features are distinct from tactile features that are derived instead from the tactile image.) \textcolor{black}{Contact} frames for surface and edge object features are shown in~\autoref{fig2} with the coordinates used to specify the poses (Equations~\ref{eq:4},\ref{eq:5} below). 

The choice of object feature has two roles. First, it defines the task: here servoing along an edge or over a surface. Second, it determines the tactile perception needed to estimate the sensor pose in the \textcolor{black}{contact} frame \textcolor{black}{${}^FP_S(t)$ from tactile measurements $\bm m(t)$ at time $t$. In general, these tactile measurements comprise a multi-dimensional time series resulting from the sensor deformation. For optical tactile sensors, such as the one used here, the values are the intensities of the pixels in the tactile image. From the tactile image, a PoseNet convolutional neural network can then predict the sensor pose in the \textcolor{black}{contact} frame of the object feature (details in \autoref{sec:3b}).} 

\subsubsection{Sensor pose}
\begin{figure*}[t!]
	\centering
	\includegraphics[width=2\columnwidth,trim={0 7 0 7},clip]{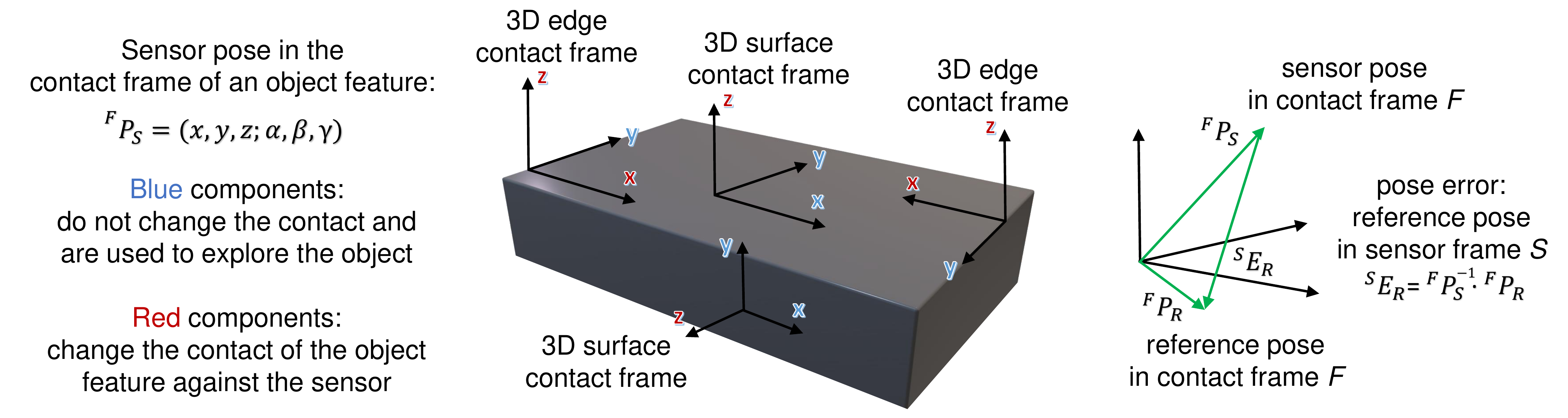} 
	\caption{Sensor poses in \textcolor{black}{contact} frames for surface or edge object features, showing only the translational axes. Motions that displace the sensor onto or away from an object feature are shown in {\textcolor{red}{red}} and invariances that keep the sensor on an object feature are {\textcolor{blue}{blue}} (assuming straight edges and flat surfaces). The pose error is the resultant motion from the sensor pose to the reference pose in the contact frame (equivalently, the reference pose in the sensor frame).}
	\label{fig2}
	\vspace{-.0em}
\end{figure*}

The geometry of these object features restricts which 3D components of the sensor pose can be predicted from the tactile measurements. For an edge parallel to the $y$-axis, or a surface parallel to the $(x,y)$-plane of the sensor frame, the 3D pose components that can be unambiguously predicted in the \textcolor{black}{contact} frame are given by the non-zero pose components as follows (\autoref{fig2}, {\textcolor{red}{red}} arrows):
\begin{align}
&\textrm{3D surface:}&{}^FP_S(t)&=(\textcolor{blue}{0},\textcolor{blue}{0},\textcolor{red}{z};\textcolor{red}{\alpha},\textcolor{red}{\beta},\textcolor{blue}{0}),\label{eq:4}\\
&\textrm{3D edge:}&{}^FP_S(t)&=(\textcolor{red}{x},\textcolor{blue}{0},\textcolor{red}{z};\textcolor{red}{\alpha},\textcolor{red}{\beta},\textcolor{red}{\gamma}).\label{eq:5}
\end{align}
Components that cannot be predicted from tactile measurements are set to zero (\autoref{fig2}, {\textcolor{blue}{blue}} arrows). These motions leave the object feature invariant (such as along an edge). 

In practice, these sensory invariances are idealizations that hold only for infinite straight edges or planar surfaces. Here we consider them to hold approximately on local regions of a curved edge or surface. In principle, this approximation could be improved with prior knowledge of the edge or surface geometry, which is interesting topic for further research. 

\subsubsection{Reference pose}
The reference in the pose error (\autoref{eq:1}) depends on the geometry of the object feature, with components in the \textcolor{black}{contact} frame:
\begin{align}
&\textrm{3D surface:}&{}^FP_R&=(\textcolor{blue}{r_x},\textcolor{blue}{r_y},\textcolor{red}{r_z};\textcolor{red}{r_\alpha},\textcolor{red}{r_\beta},\textcolor{blue}{r_\gamma}),\label{eq:7}\\
&\textrm{3D edge:}&{}^FP_R&=(\textcolor{red}{r_x},\textcolor{blue}{r_y},\textcolor{red}{r_z};\textcolor{red}{r_\alpha},\textcolor{red}{r_\beta},\textcolor{red}{r_\gamma}).\label{eq:8}
\end{align} 
The \textcolor{blue}{blue} components of the reference will set how the object is explored and the \textcolor{red}{red} components will set the desired contact. 

By comparing the predicted sensor pose~(Equations~\ref{eq:4},\ref{eq:5}) with the reference (Equations~\ref{eq:7},\ref{eq:8}), we see two types of pose error component that differ in their actions on the sensor pose:\\
\noindent (i) those where both the reference and predicted sensor~pose can be non-zero, which change how the tactile~sensor contacts the object feature (colored {\textcolor{red}{red}} in Figure~\ref{fig2} and \mbox{Equations~\ref{eq:4}-\ref{eq:8}});\\ 
\noindent (ii) those where just the reference can be non-zero, which are invariances that do not change how the sensor contacts the object feature (colored {\textcolor{blue}{blue}} in Figure~\ref{fig2} and Equations~\ref{eq:4}-\ref{eq:8}). These {\textcolor{blue}{blue}} reference components move the sensor in a way that the contact frame appears identical to the sensor. We use these motions for exploration steps along the edge or surface. 


In practice, these displacements and invariances hold approximately on local regions of an object. As the sensor moves over a curved edge or surface, the sensor pose is disturbed in the \textcolor{black}{contact} frame on a local part of the object (which itself moves relative to the fixed base frame of the robot when the object is non-uniform). The disturbance produces a pose error, which drives the controller to move the sensor towards the reference in the new \textcolor{black}{contact} frame. Repeating this procedure servos the tactile sensor over a curved edge or surface. 

Note that the geometry of the object feature determines how a tactile sensor can explore that feature extended over the object. This is different from Li {\em et al}'s approach (see~\autoref{sec:2a}), which composed an external motion signal with the controlled motion~\cite{li_control_2013}. In principle, the reference components for the invariances could be set to zero (\mbox{Equations~\ref{eq:7},\ref{eq:8}}, \textcolor{blue}{blue} terms) and the control composed with an external signal. However, an external motion signal can disturb the pose components that are being controlled. Therefore, we unify the two motions within a single reference, keeping complete freedom to explore the object without disturbing the servo control.

\subsubsection{Controller}
It remains to specify the control law that aims to drive the pose error in \autoref{eq:1} to zero. In this work, we control the sensor pose using a proportional-integral (PI) controller in discrete time ($t=0,1,2,\ldots$). The controller computes a correction to the sensor pose, ${}^{S}U_{S'}(t)=U(t)$, from the pose error, ${}^SE_R(t)=E(t)$, using a parameterised representation in Cartesian coordinates and Euler angles:
\begin{equation}
\label{eq:9}
U(t) = K_{\rm P}\,E(t) + K_{\rm I}\,f\left[\sum_{\tau=0}^{t}\,E(\tau)\right].
\end{equation}
For the purpose of specifying the controller, these parameterised ${\rm SE(3)}$ transformations are treated as 6D vectors $U=(u_x,u_y,u_z;u_\alpha,u_\beta,u_\gamma)$ and $E=(e_x,e_y,e_z;e_\alpha,e_\beta,e_\gamma)$ in the coordinates of the sensor frame. The $6\times6$ gain matrices $K_P$ and~$K_I$ act on these 6D vectors, and are supplemented with an anti-windup function $f[\cdot]$ to bound the integral error between a minimum and maximum. 

 
In our system, position control of the robot is specified in its base frame $B$, which is set at the start of the experiment along with an initial sensor pose ${}^BP_S(0)$. Then the control signal ${}^{S}U_{S'}(t)$ updates the sensor pose in the base frame:
\begin{equation}
{}^BP_{S}(t+1)={}^BP_{S'}(t)={}^BP_S(t) {}^{S}U_{S'}(t).
\end{equation}

\subsubsection{Implementation}\label{sec:3a6}
All poses $(x,y,z;\alpha,\beta,\gamma)\in{\rm SE}(3)$ are represented in Cartesian coordinates and Euler angles (in the extrinsic-$xyz$ convention). 
The sensor pose components in the \textcolor{black}{contact} frame (Equations~\mbox{\ref{eq:7},\ref{eq:8}}) are assumed to lie within ranges: $x\in[-5,5]\,$mm horizontally centred on the edge; $z\in[-5,-1]$\,mm vertically into the surface or top of the edge; $\alpha,\beta\in[-15^\circ,15^\circ]$ roll and pitch around the surface/edge normal; and $\gamma\in[-45^\circ,45^\circ]$ yaw centred on the edge (Appendix:~\autoref{tab:2}). These ranges determine the span of data used to train the neural network for predicting the sensor pose. A typical reference pose is set in the center of these ranges, representing a $-3\,$mm contact depth with the sensor oriented normal to the surface or edge. 

The control gain matrices are diagonal and have values that give good overall performance while not being specific to any of our experiments (Appendix: \autoref{tab:1}). Proportional gains are $0.5$ for components of the displacement ({\textcolor{red}{red}}) and $1$ for invariances ({\textcolor{blue}{blue}}); integral gains are $0.3$ for translational and $0.1$ for rotational displacements, and $0$ for the invariances. The integrated error bounds are $\pm5\,$mm for translations, $\pm15^\circ$ for roll and pitch, and $\pm45^\circ$ for yaw rotations.

\section{Pose prediction from deep learning}\label{sec:3b}

Pose-based tactile servo control requires an estimate of the sensor pose {${}^FP_S(t)$} in the \textcolor{black}{contact} frame of an edge or surface feature, from tactile measurements $\bm m(t)$ at time $t$. In related work~\cite{lepora_optimal_2020}, we investigated how deep learning can be used to train accurate `PoseNet' neural network models for estimating 3D pose from tactile images. This involved a systematic approach for selecting the best network hyperparameters, using Bayesian optimization applied to the validation loss, rather than the more common approach of hand-tuning. 

This work uses the methods in \cite{lepora_optimal_2020} to gather training/validation data and train the neural network. We emphasise that the data collection involved a transverse sliding motion prior to recording each tactile image via an (unlabelled) random displacement of the sensor. This motion is important because contact-dependent shear affects measurements from optical tactile sensors such as the one used here~\cite{aquilina_shear-invariant_2019,lepora_pixels_2019}. 

We refer to \cite{lepora_optimal_2020} for the details of the methods, and report the differences from that previous analysis. These were: (i)~larger training/validation datasets of 5000/5000 samples over the same ranges as previously~(Appendix: \autoref{tab:2}); (ii)~fewer iterations (100) for the hyperparameter optimization by fixing values that were consistently optimized~\cite{lepora_optimal_2020}, such as having 5 convolutional and 1 dense layer~(Appendix: \autoref{tab:3}). 

The accuracies of these surface and edge PoseNets are reported below (\autoref{tab0}). The horizontal and yaw pose accuracies were much improved over previous results~\cite{lepora_optimal_2020}; other pose parameters have similar or slightly poorer accuracies, which we attribute to unimportant differences in the sensor and its placement during data collection. The key point is that this method gives accurate pose estimates that are insensitive to shear as the sensor interacts with an object.

\begin{table}[h]
	\begin{tabular}{c|cc|cc}
		\textbf{Parameter} & \textbf{Surface (MAE)} & \textbf{Range} & \textbf{Edge (MAE)} & \textbf{Range} \\ 
		\hline
		horizontal, $x$ & - & - & 0.3\,mm & 10\,mm\\ 
		vertical, $z$ & 0.1\,mm & 4\,mm & 0.2\,mm & 4\,mm \\
		roll, $\alpha$ & $0.4^\circ$ & $30^\circ$ & $1.2^\circ$ & $30^\circ$ \\ 
		pitch, $\beta$ & $0.5^\circ$ & $30^\circ$  & $2.4^\circ$ & $30^\circ$ \\ 
		yaw, $\gamma$ & -& - &$4.1^\circ$ & $90^\circ$ \\ 
	\end{tabular}
	\caption{PoseNet accuracy: Mean Absolute Error (MAE) of predictions. Tests were performed over an independent set of 2000 samples.}
	\vspace{-2em}
	\label{tab0}
\end{table}

\section{Tactile robotic system}\label{sec4}

\subsection{Tactile sensor}\label{sec:4a1}

This study involves the use of a soft biomimetic optical tactile sensor from the BRL TacTip family~\cite{ward-cherrier_tactip_2018,chorley_development_2009}. The use of 3D-printing enables the designs to be highly customizable, encompassing stand-alone sensors and fingertips of robot hands with various shapes and sizes~\cite{ward-cherrier_tactip_2018}. \textcolor{black}{Printing multiple materials in one piece allows combination of a rigid casing (VeroWhite) with a soft sensing surface (TangoBlack).} The version used here has a soft tactile dome (40\,mm dia.) with 127 internal pins tipped by printed white markers that function as tactile elements.

An important aspect of the soft biomimetic design is that the sensor is highly sensitive to shear deformation, which induces a global movement of all pins along the direction of lateral strain in the sensing surface. While this is useful for detecting phenomena such as slip~\cite{james_slip_2020}, it causes issues for tactile servoing when sliding over object surfaces~\cite{aquilina_shear-invariant_2019,lepora_pixels_2019}. The problem is that the tactile image will depend not only on the pose of the object but also on the prior transverse movement of the sensor. As shown in previous work~\cite{lepora_pixels_2019}, this causes the tactile servoing to fail under sliding motion unless the pose can be inferred accurately. Here we mimic the effect of this transverse motion during training data collection, so that the PoseNet is trained to be insensitive to sensor motion. 


\begin{figure}[t!]
	\centering
	\includegraphics[width=0.585\columnwidth,trim={140 100 140 40},clip]{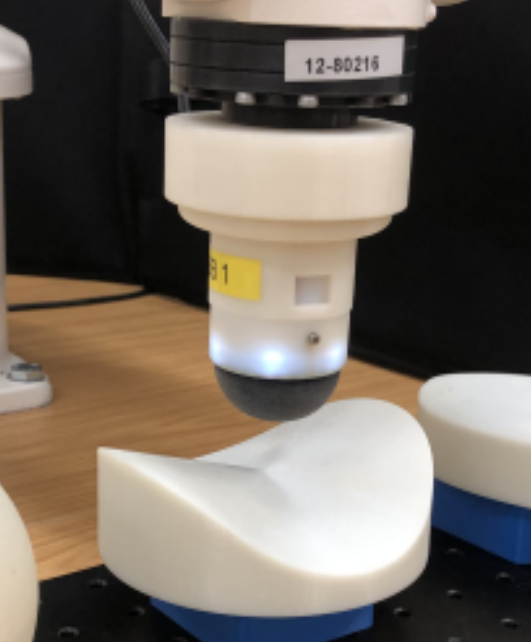}
	\includegraphics[width=0.4\columnwidth,trim={25 5 70 20},clip]{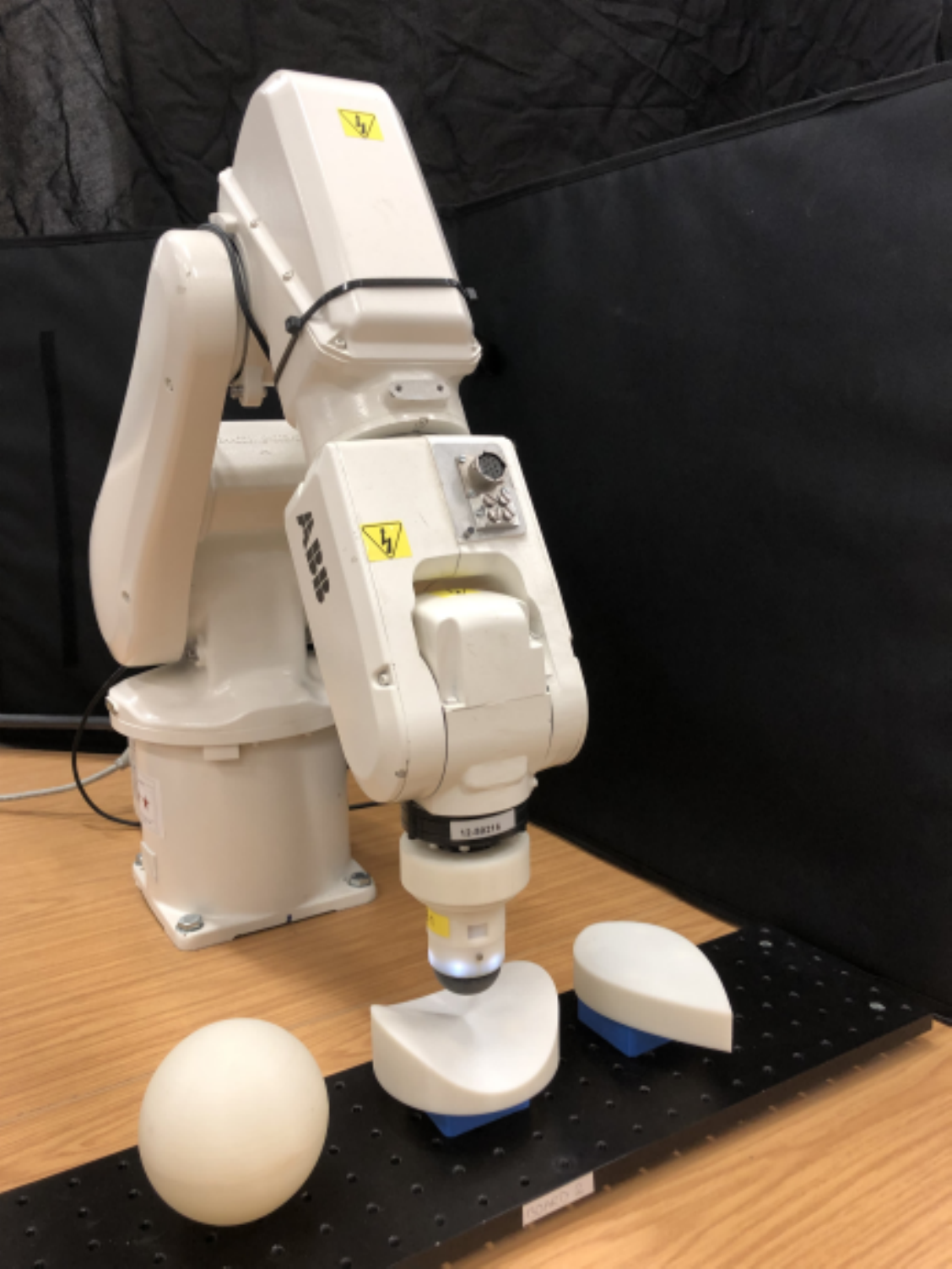}
	\caption{Tactile robotic system, showing (left) a close-up of the tactile sensor and (right) a view of the sensor-arm system next to some test objects.}
	\label{fig6}
	\vspace{-0.5em}
\end{figure}

\subsection{Tactile robot}\label{sec:4a2}

The TacTip optical tactile sensor is mounted as an end-effector on an industrial robot arm (\autoref{fig6}), which was also used in earlier studies of control using this tactile sensor~\cite{lepora_exploratory_2017,aquilina_shear-invariant_2019,lepora_pixels_2019}. We use an IRB~120 (ABB Robotics): a fairly small 6-axis industrial robot arm (reach $0.58\,$m, $25\,$kg weight, $3\,$kg payload) with accurate positioning ($0.01\,$mm accuracy, \mbox{$\sim 1$\,sec} cycle time). The TacTip base is bolted onto a mounting plate attached to the rotating (wrist) section of the arm.

Closed-loop control of this robot is limited to long latencies ($\gtrsim$ $100\,$ms), since the IRB 120 is aimed at applications where its end-effector passes through predefined waypoint poses. Therefore, we control the arm in discrete updates of the end-effector pose, with the industrial robot controller (IRC5) calculating the motion between poses. While this setup is primitive compared with many research robots, {\em e.g.} cobots that react to humans in real-time, it does provide a basic platform sufficient for investigating pose-based tactile servo control. 

\begin{figure*}[t!]
	\centering
	\begin{tabular}[b]{@{}c@{}}
		\textbf{(a) Pose-based tactile servo control: 3D sphere} \\
		\includegraphics[width=0.8\columnwidth,trim={40 15 40 30},clip]{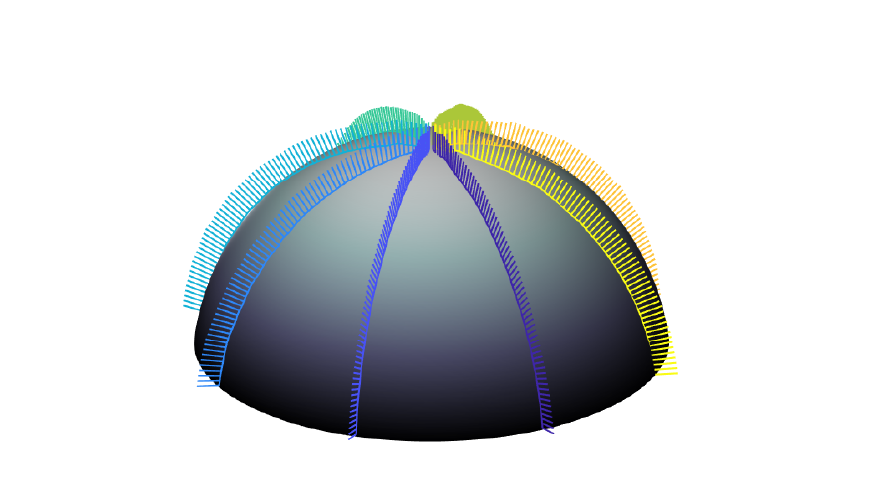} 
		\includegraphics[width=0.8\columnwidth,trim={80 100 50 45},clip]{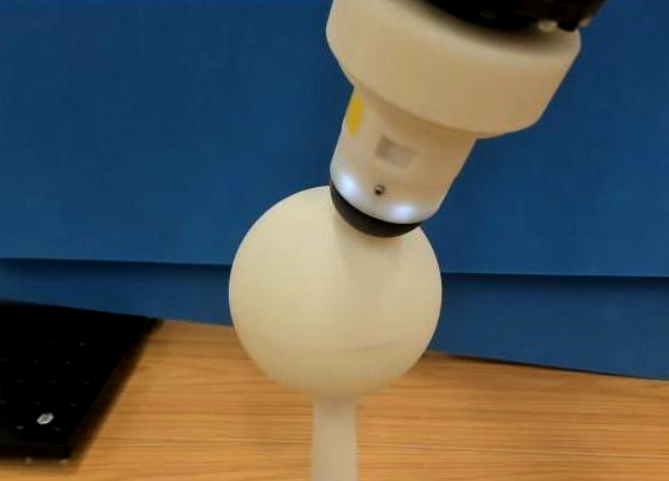} \\
		\begin{tabular}[b]{@{}c@{}c@{}}
			\begin{tabular}[b]{@{}c@{}}
				\textbf{(b) PoseNet predictions} \\
				\includegraphics[width=1.15\columnwidth,trim={60 0 40 0},clip]{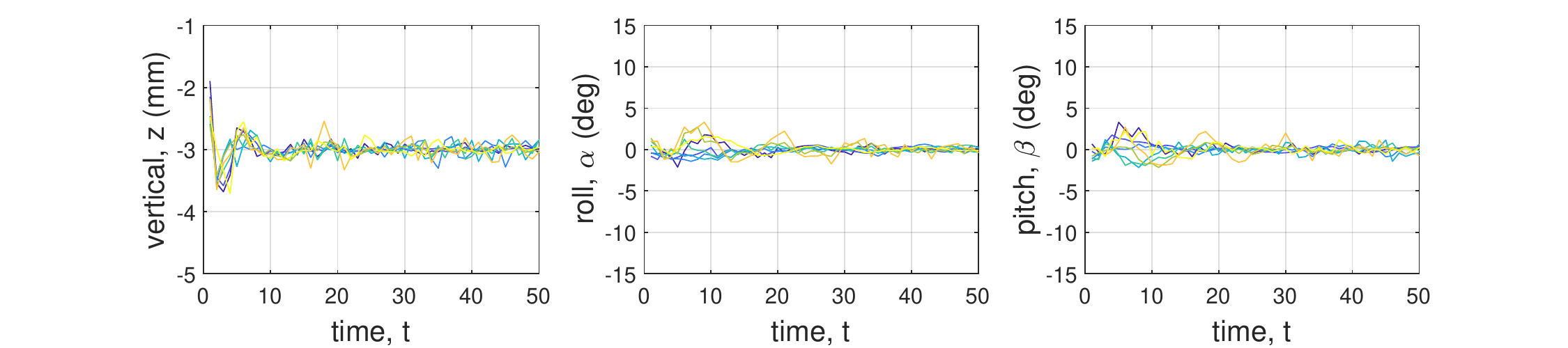} \\
				\textbf{(c) Deviations from sphere surface} \\
				\includegraphics[width=1.15\columnwidth,trim={60 0 40 0},clip]{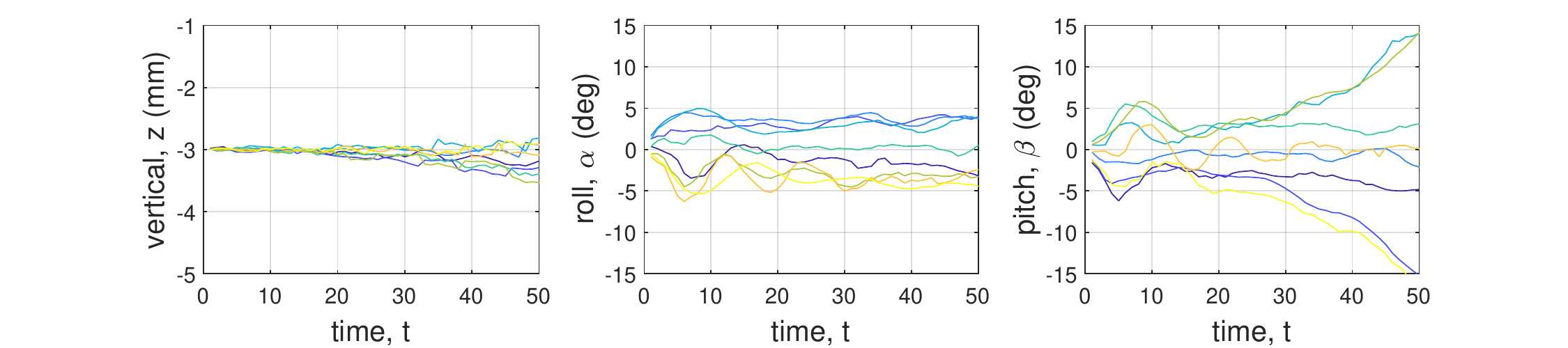} \\
			\end{tabular} & 
			\begin{tabular}[b]{@{}c@{}}
				\textbf{(d) Tactile servo control: Yaw set pose} \\
				\includegraphics[width=0.75\columnwidth,trim={60 0 38 15},clip]{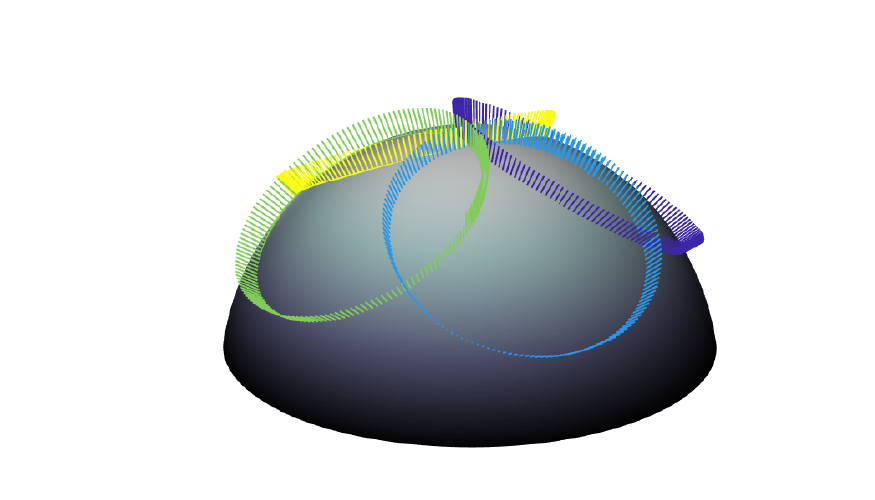} \\
			\end{tabular}
		\end{tabular}
	\end{tabular}
	\caption{Pose-based servo control over a sphere. (a) Trajectories with predicted surface normals ($z$-axis) are shown for 8 reference poses with angles $(r_x,r_y)=(\cos\varphi,\sin\varphi)$ every $45^\circ$. The corresponding PoseNet predictions (b) and deviations from the surface (c) are plotted against time. (d) Trajectories when the yaw reference is non-zero, $r_\gamma=2^\circ$, at angles $\varphi=0^\circ,90^\circ,180^\circ,270^\circ$. A video for the 8 reference poses is in \href{https://youtu.be/PrtpVU-a1rk}{Supplementary Movie S1}}.\\
	\label{fig11}
	\vspace{1em}
	\begin{tabular}[b]{@{}c@{}}
		\textbf{(a) Pose-based tactile servo control: 3D disk} \\
		\includegraphics[width=0.8\columnwidth,trim={30 75 30 80},clip]{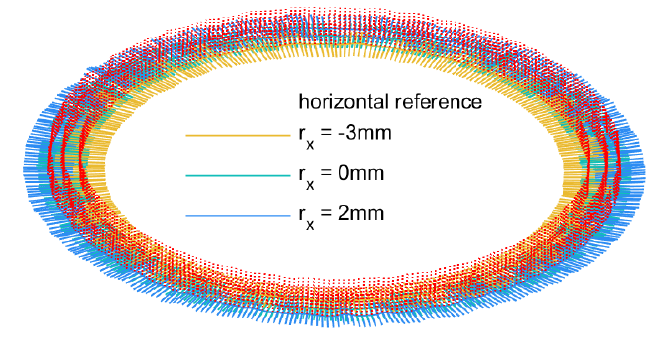} 
		\includegraphics[width=0.8\columnwidth,trim={50 20 50 20},clip]{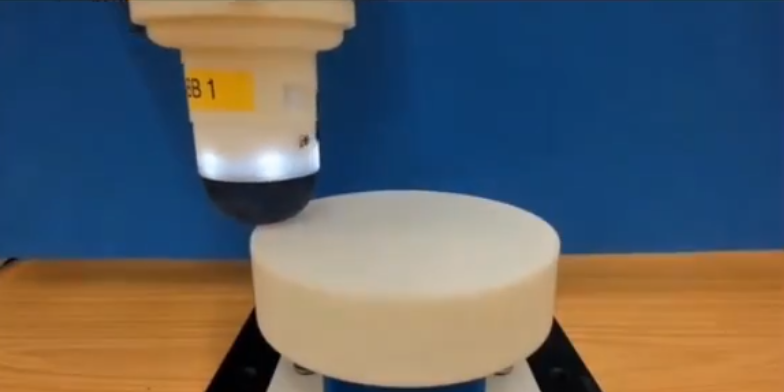} \\
		\textbf{(b) PoseNet predictions} \\
		\includegraphics[width=1.75\columnwidth,trim={100 0 80 0},clip]{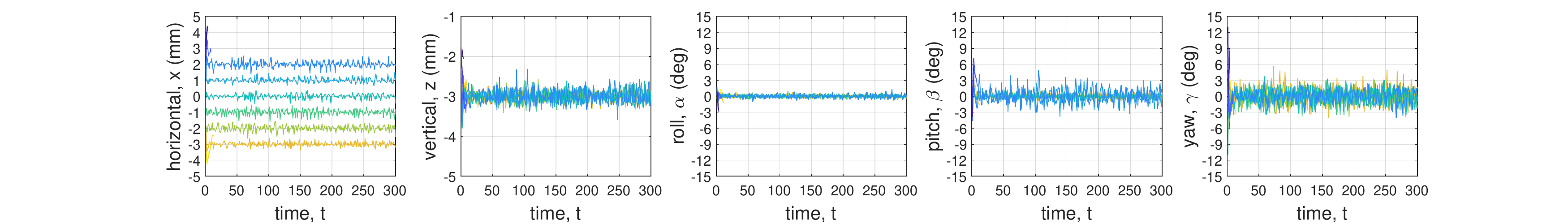} \\
		\textbf{(c) Deviations from disk edge} \\
		\includegraphics[width=1.75\columnwidth,trim={100 0 80 0},clip]{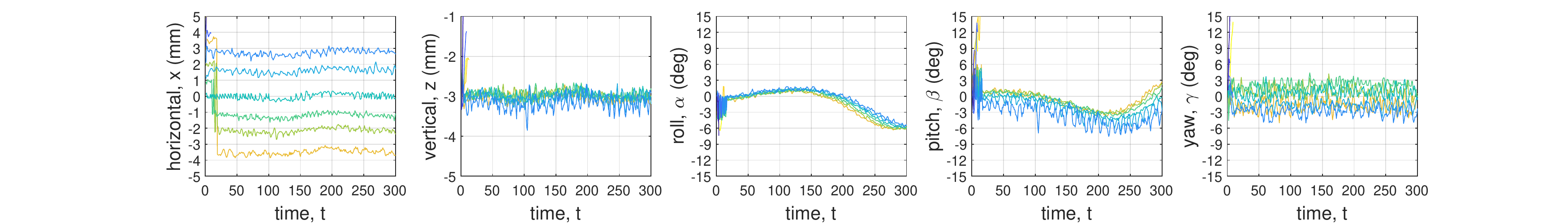} \\
	\end{tabular}
	\vspace{-0.5em}
	\caption{Pose-based servo control around a disk edge. (a) Trajectories for 3 reference poses $r_x=-3,0,3\,$mm with predicted surface normals and edge normals, (b) PoseNet predictions and (c) deviations from the disk edge for reference poses over $[-3,2]\,$mm. A video is in \href{https://youtu.be/UXIM8k3Pjqs}{Supplementary Movie S2}.}
	\label{fig12}
\end{figure*}

\subsection{Software infrastructure}

The software and hardware are integrated with four main components written mainly in Python: (1) Tactile images are collected using the OpenCV library, then preprocessed and saved for training, or used directly for prediction; (2)~These images are cropped, thresholded and subsampled in OpenCV to give $(128\times128)$-pixel images, which are then passed to the deep learning models implemented using the Keras deep learning library; (3) The resulting predictions are passed to control software in Python (with some visualizations using the MATLAB Engine API); (4)~Finally, the computed control signal is sent to the robot arm via the RAPID API. 

Training and optimization of the deep neural networks was implemented on a Titan Xp GPU hosted on a Windows 10~PC. A training run typically takes about 10 minutes. The learnt model is then copied to the robot control PC, which in this case was a Windows 10 PC with limited GPU hardware. Even so, the PoseNet models typically made predictions in about $50\,$msec. The longest delay within the cycle was the latency between the control PC and the robot arm (typically $100\,$ms). 

\section{Tactile servo control on regular objects}\label{sec:6}

\begin{figure*}[t]
	\centering
	\begin{tabular}[b]{@{}c@{}}
		\textbf{Mean deviations from 3D spherical surface, varying reference pose} \\
		\includegraphics[width=2\columnwidth,trim={100 5 50 10},clip]{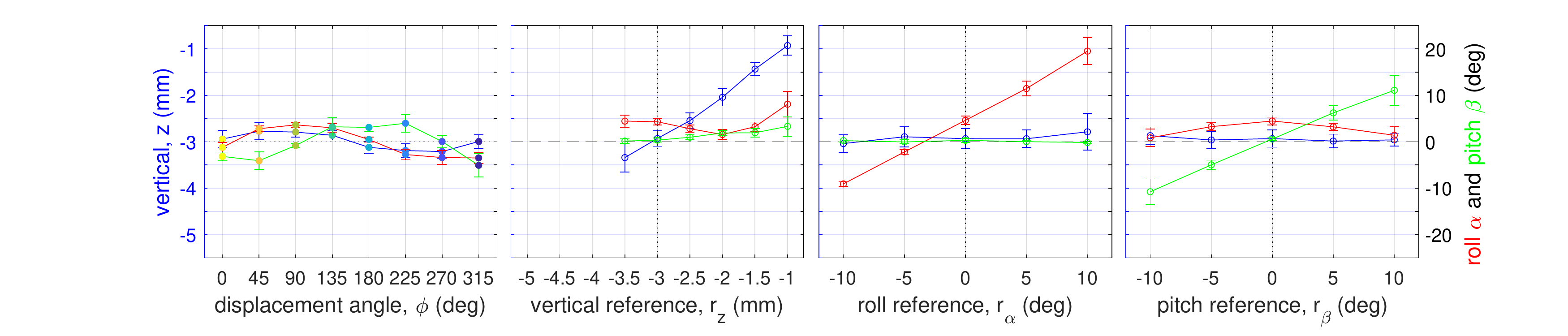} \\
	\end{tabular}
	\vspace{-0.5em}
	\caption{Effect of changing the reference pose on PBTS control for the spherical surface. The mean deviations and spreads (1 s.d.) of the pose deviations are calculated over a trajectory over the surface of a sphere, with deviations colored according to their pose component (shown on the left and right axes). The colored markers for $\varphi$ correspond to the trajectories in \autoref{fig11} (panels a-c).}
	\vspace{0em}
	\label{fig10}
	\vspace{1em}
	\centering
	\begin{tabular}[b]{@{}c@{}}
		\textbf{Mean deviations from 3D disk edge, varying reference pose} \\
		\includegraphics[width=2\columnwidth,trim={80 5 50 10},clip]{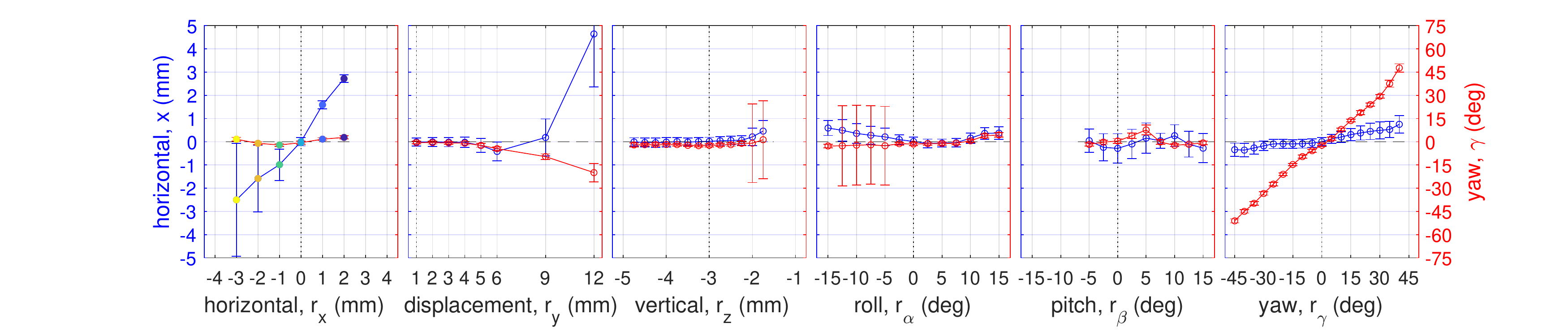} \\
		\includegraphics[width=2\columnwidth,trim={80 5 50 10},clip]{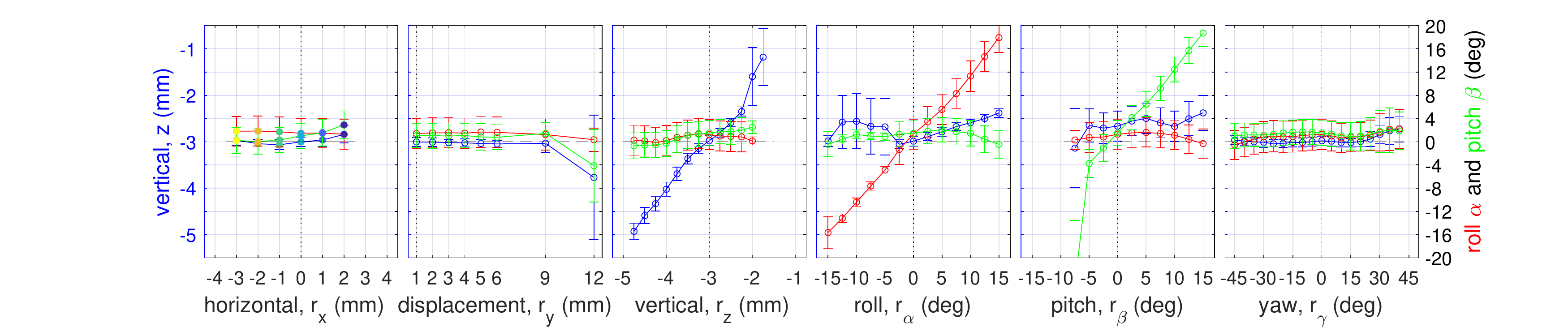} \\
	\end{tabular}
	\vspace{-0.5em}
	\caption{Effect of changing the reference pose on PBTS control for the 3D disk, showing (a) the horizontal and yaw deviations; and (b) the vertical, roll and pitch deviation. The mean deviations and spreads ($\pm1$ s.d.) are calculated over a trajectory around the disk, with deviations colored according to their pose component (shown on the axes). The colored markers for $r_x$ correspond to the trajectories in \autoref{fig12}.}
	\vspace{-0em}
	\label{fig13}
\end{figure*}

\subsection{3D spherical surface}\label{sec:6b}

\begin{figure*}[t!]
	\centering
	\begin{tabular}[b]{@{}c@{}c@{}c@{}}
		\textbf{3D saddle surface} & \textbf{3D wave surface} & \textbf{Head surface}\\
		\includegraphics[width=0.8\columnwidth,trim={20 0 10 30},clip]{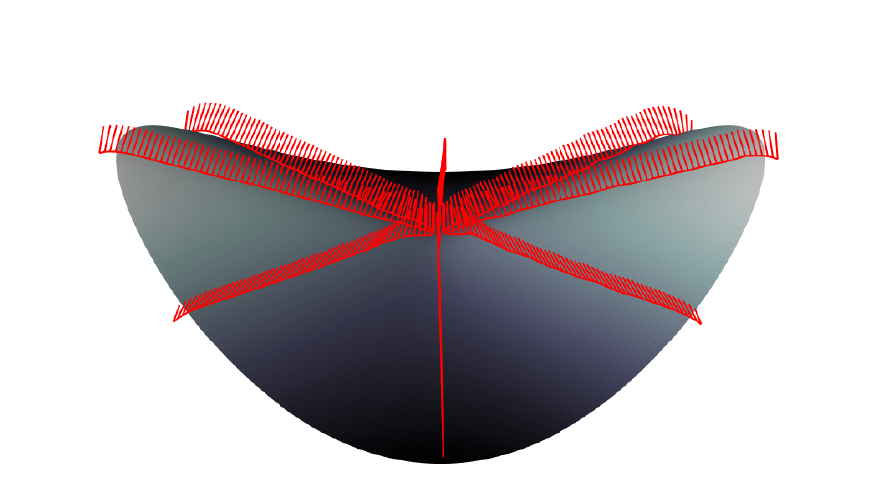} & 	
		\includegraphics[width=0.75\columnwidth,trim={10 -10 10 0},clip]{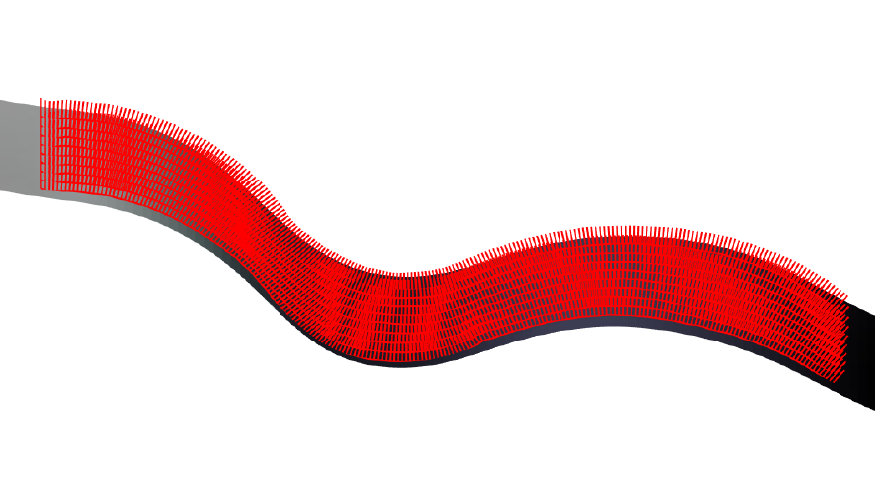} &
		\includegraphics[width=0.66\columnwidth,trim={60 10 30 5},clip]{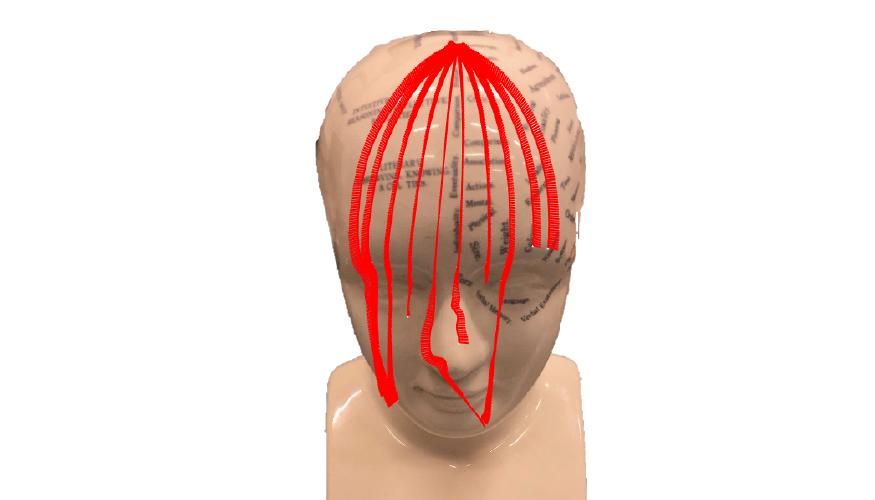} \\
	\end{tabular}
	\begin{tabular}[b]{@{}c@{}c@{}c@{}c@{}}
		\textbf{Planar volute} &
		\textbf{Planar teardrop} &
		\textbf{Planar clover} &
		\textbf{Planar spiral} \\
		\includegraphics[width=0.5\columnwidth,trim={20 20 20 20},clip]{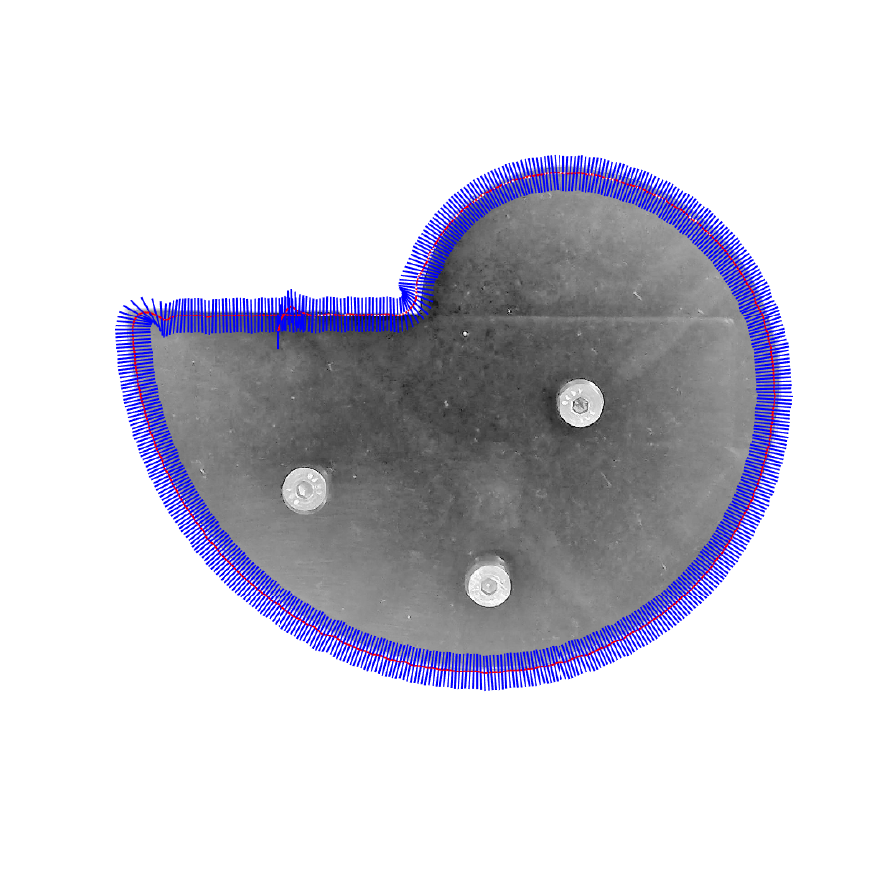} &
		\includegraphics[width=0.5\columnwidth,trim={20 20 20 20},clip]{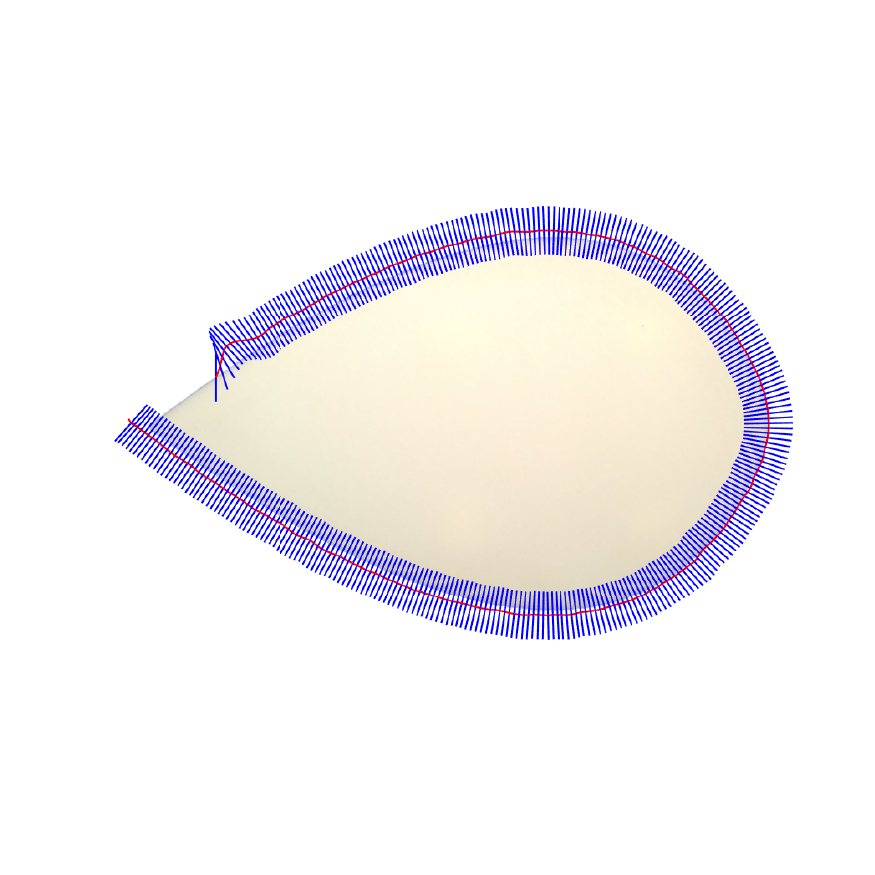} &		
		\includegraphics[width=0.5\columnwidth,trim={20 20 20 20},clip]{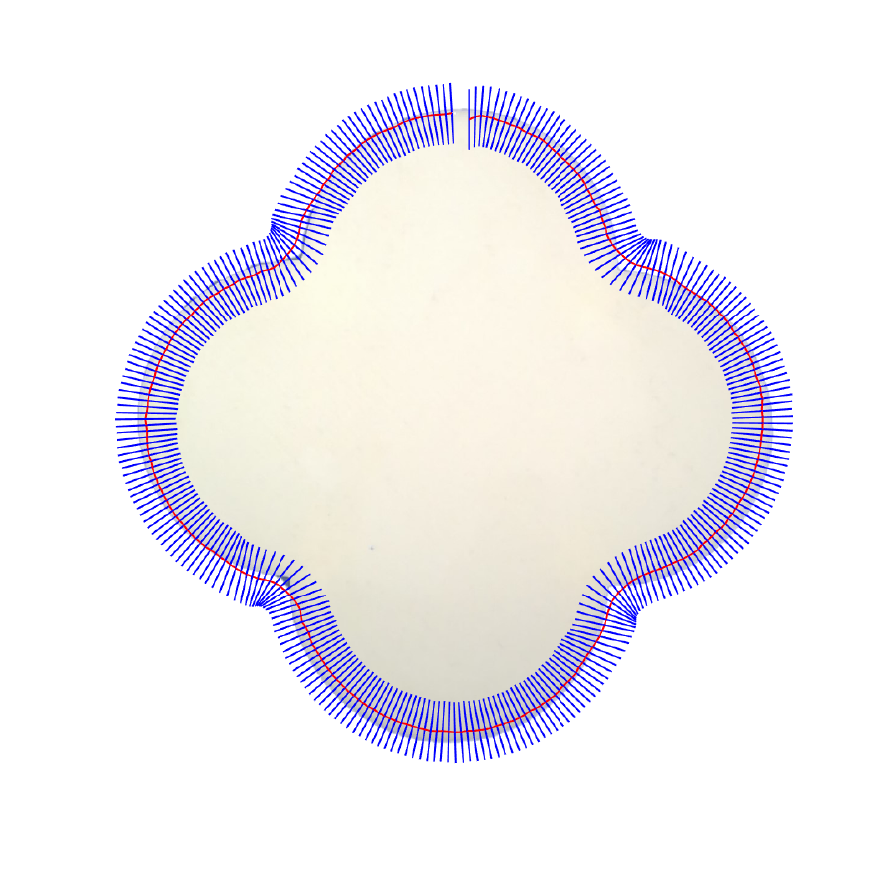} &
		\includegraphics[width=0.5\columnwidth,trim={20 20 20 20},clip]{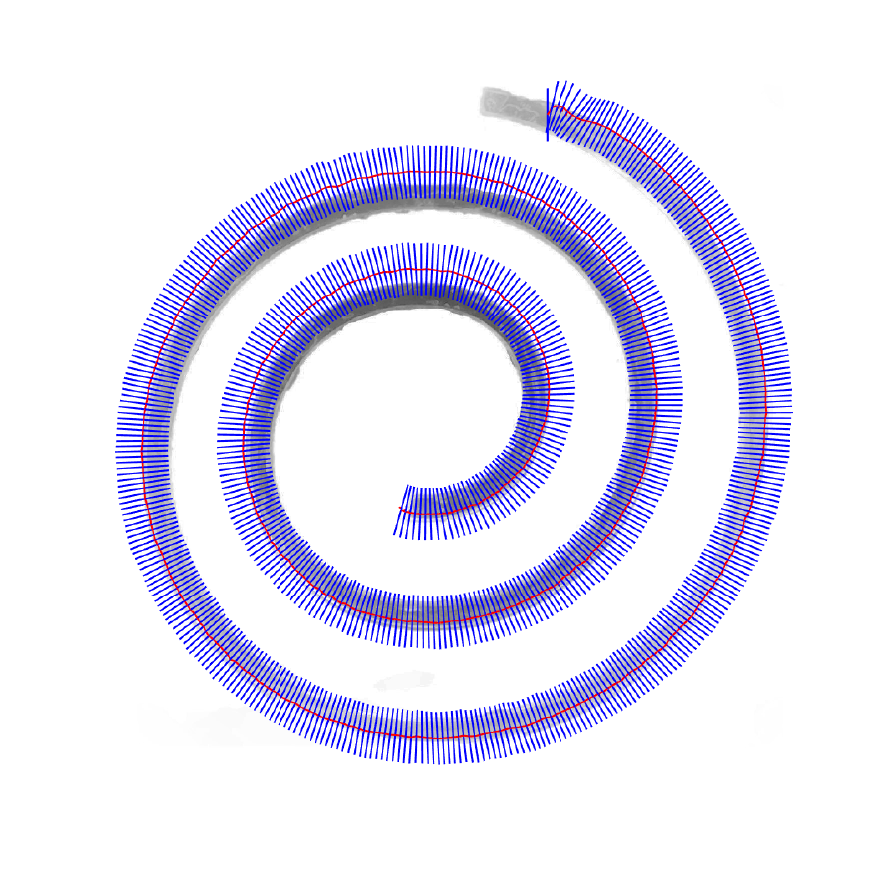} \\
	\end{tabular}
	\vspace{-.5em}
    \begin{tabular}[b]{@{}c@{}c@{}c@{}}
		\textbf{3D saddle edge} & \textbf{3D wave edge} & \textbf{Lid edge} \\
		\includegraphics[width=0.8\columnwidth,trim={20 0 10 25},clip]{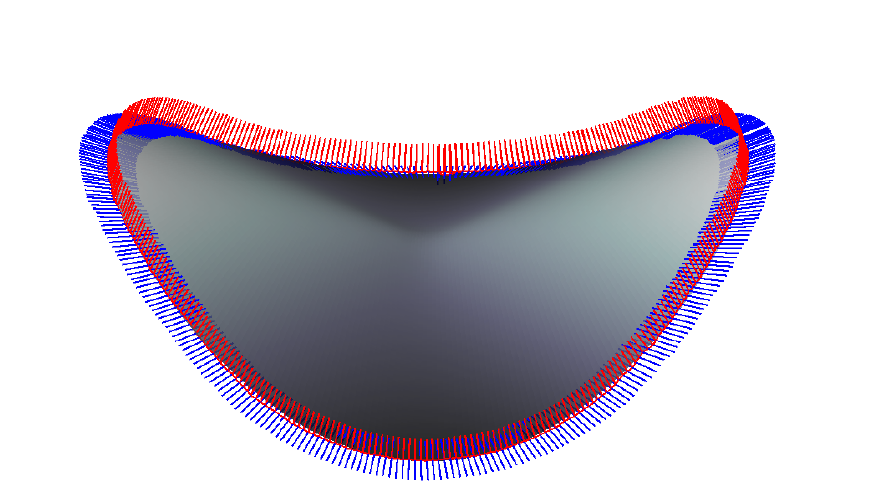} &
		\includegraphics[width=0.75\columnwidth,trim={0 -10 0 0},clip]{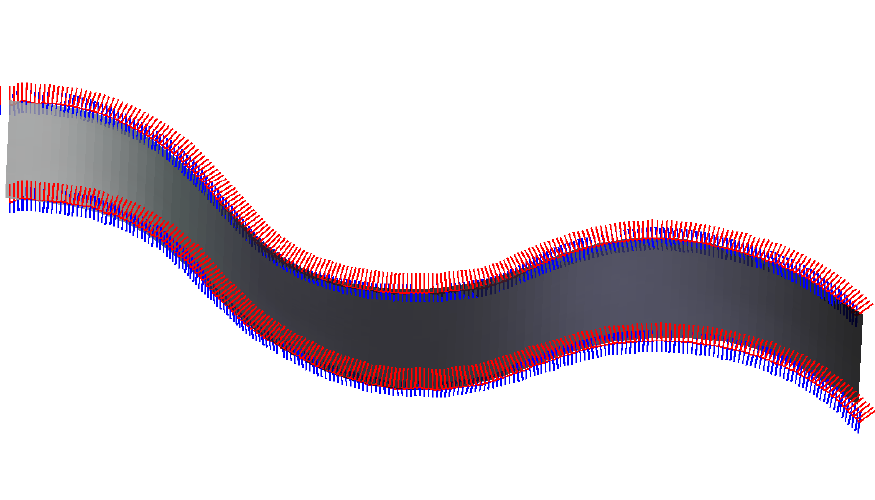} & 
		\includegraphics[width=0.6\columnwidth,trim={55 0 30 15},clip]{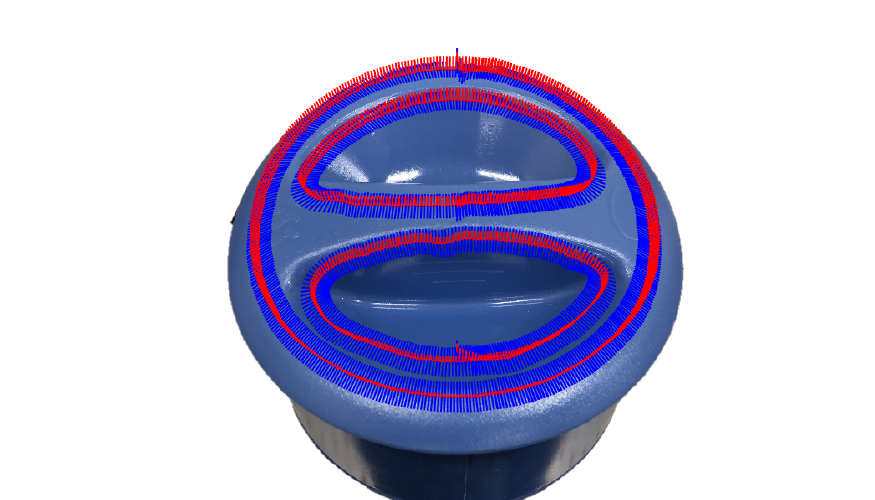} \\
	\end{tabular}
	\caption{Pose-based control over irregular objects, following 3D surfaces (top row), planar edges (middle row) and 3D contours (bottom row). Predicted surface normals ($z$-axis, red) and edge normals ($x$-axis, blue) are shown. All objects were novel because the PoseNet was trained on the disk. Supplementary videos are in \href{https://youtu.be/NldNjDKNS-I}{Movie S3} (planar edged objects), \href{https://youtu.be/7CfnEZGgBRc}{Movie S5} (wave edge and surface), \href{https://youtu.be/QC3Q6ArYoUE}{Movie S6} (saddle edge and surface), \href{https://youtu.be/N98nVqXvm88}{Movie S7} (head) and \href{https://youtu.be/x-xnLj_RgAk}{Movie S8} (lid).}
	\vspace{-.5em}
	\label{fig14}
	\vspace{-.5em}
\end{figure*}

First, the pose-based servo control was tested on a 3D spherical surface, starting at the apex and moving outwards and downwards. \textcolor{black}{For the tactile sensor used here, a reference depth of $-3\,$mm normal to the surface gave reliable contacts without being close to damaging the sensor.} An exploration step size of $1\,$mm in a direction at angle $\varphi$ was set with a reference pose ${}^FP_R=(\cos\varphi,\sin\varphi,-3;0,0,0)$.

The resulting trajectories followed the sphere surface accurately after initially aligning the sensor to the surface (\autoref{fig11}a, colored by angle~$\varphi=0^\circ,45^\circ,\cdots315^\circ$). The PoseNet predictions match the estimated deviations with a small bias of a few degrees in the roll and pitch deviations and a vertical deviation of less than $0.1\,$mm for the first 30 mm-steps, then drifting up to around $0.4\,$mm and a large deviation in pitch (\autoref{fig11}b,c). From inspection, the sensor measurements seem to depend on gravity, with the markers moving downwards as the tip is oriented horizontally. We therefore attribute this drift to gravity deforming the sensor tip as it orients horizontally relative to the more vertical pose during training.

We also considered servo control with reference poses ${}^FP_R=(\cos\varphi,\sin\varphi,-3;0,0,2)$ that rotate the sensor in $\gamma=2^\circ$ steps while moving over the surface (\autoref{fig11}d). The accuracy is visible in the way that the trajectories close at the apex of the sphere rather than drift away from perfect circles.

Performance over each trajectory was summarized by the mean and spread of the vertical, roll and pitch $(z,\alpha,\beta)$ deviations from the normal to the surface (\autoref{fig10}). These deviations from the sphere, although small, appear periodic in the angle $\varphi$ that sets the direction of the trajectory (\autoref{fig10}b, left panel). This periodicity is consistent with the effect of gravity deforming the sensor, since the trajectory angle determines which part of the sensing surface points upwards and hence the orientation of the roll and pitch.

The servo control was then tested over reference poses ${}^FP_R=(1,0,r_z;r_\alpha,r_\beta,0)$ ranging in contact depth, roll and pitch. Overall, the servoing remained accurate, with small vertical deviations of $\lesssim0.2\,$mm and small angular deviations of $(\alpha,\beta)\lesssim(5^\circ,0^\circ)$. The deviation also varied linearly with its corresponding reference component (\autoref{fig10}b, right panels). Overall, the servo control has the expected effect on the deviations of the trajectories from the true surface normals. 

\subsection{3D disk edge}\label{sec:6c}
 
Pose-based servo control was then tested on a planar disk. \textcolor{black}{Again, a contact depth of $-3\,$mm normal to the disk was used.} An exploration step size of $1\,$mm along the edge was set with a reference pose ${}^FP_R=(0,1,-3;0,0,0)$.

The resulting trajectory precisely followed the outside of the disk after an initial motion to align the sensor to the edge (\autoref{fig12}, green trajectory). The PoseNet predictions match the estimated deviations from the edge to $\lesssim0.3\,$mm horizontally, $\lesssim0.2\,$mm vertically and $\lesssim3^\circ$ yaw. There is an apparent periodic deviation in the roll and pitch (\autoref{fig12}c). An inspection revealed the sensor was mounted slightly off-normal to the wrist, which caused it to precess as the wrist rotated. This inaccurate mounting was corrected by the servo control to give an apparent deviation in roll and pitch.  

Tactile servoing around the 3D disk was then examined over a range of reference poses ${}^FP_R=(r_x,r_y,r_z;r_\alpha,r_\beta,r_\gamma)$, with the performance over each trajectory summarized by the mean and spread of the deviation from the disk edge. The servoing remained accurate over these ranges, with variations in each reference component mainly affecting only the corresponding pose component in a linear manner (\autoref{fig13}). 

Both the 3D edge and sphere had similar deviations along the vertical, roll and pitch with changing the reference components. Vertical deviations were more accurate for the edge than the surface, even though the surface PoseNet is more accurate than the edge PoseNet. We attribute this to the effects of gravity on the sensor tip (as described in \autoref{sec:6b}). 

\section{Tactile servo control on irregular objects}\label{sec:7}

\subsection{Irregular 3D surfaces}\label{sec:7b}

Pose-based servo control was next trialled on several irregular 3D shapes: a saddle, a 3D wave and a complex surface (a model of a head), \textcolor{black}{again using a reference pose with $-3$\,mm contact depth and 1\,mm exploration step}. As the 3D PoseNet was trained on a planar surface, successful tests demonstrated its robustness to moderately curved surfaces. 

All irregular surfaces were successfully traced (\autoref{fig14}, top row). The close match of the trajectories to the the objects is seen by overlaying the motion over the design of the saddle and wave, and observing the likeness to a photographic image of the head. (Note that some paths were shorter because of a singularity in the robot's workspace on the right eye socket.)  
 
An additional test was on a soft object: a `squishy brain'. The sensor was initially placed at a location where the object was deformed, servoing to make a gentler contact while tracing the object surface. Both the object and sensor deform initially before the sensor maintains a light touch. A supplementary video showing this behavior is in \href{https://youtu.be/BWrq5z0dLf4}{Movie S9}.
 
\subsection{Irregular planar edges}\label{sec:7a}
 
The next trials were on a range of irregular planar shapes: a volute, teardrop, clover and spiral \textcolor{black}{(reference pose with $-3$\,mm contact depth and 1\,mm exploration step)}. The success of these tests demonstrated robustness to positive and negative edge curvatures, including corners. 
 
All irregular planar contours closely matched the true shapes, in comparison to the underlaid photos of the objects (\autoref{fig14}, middle row). There was a failure mode at very sharp corners where the 3D controller could not complete the teardrop point and the inward tip of the spiral. Near these points, large deviations in roll and pitch occurred that took the sensor away from the vertical (\href{https://youtu.be/NldNjDKNS-I}{Supplementary Movie S3}). 

However, the PoseNet model was only trained on edges, so any success on a corner (such as the two on the planar volute) is an on emergent property of the control. Further, for sensor motion constrained to maintain a vertical orientation, all contours were successfully traced (\href{https://youtu.be/nlFep15iZdE}{Supplementary Movie S4}), as also found in previous work on 2D servo control~\cite{lepora_pixels_2019}.


\subsection{Irregular 3D edges}

Trials were then around several irregular 3D contoured shapes: the edges of the 3D wave and saddle used as surfaces above, and a complex contoured object (a container lid), \textcolor{black}{using the same reference pose}. The success of these tests demonstrated robustness to varying edge curvatures and profiles.

All plotted contours closely followed the shapes of the irregular 3D objects (\autoref{fig14}, bottom row). This is visible in overlays of the 3D designs of the saddle and wave, and the good likeness to the image of the complex-shaped lid. We draw particular attention to the accurate tracing of the lid, which was challenging because its `edge' was much blunter than the one used in training. 




\section{Discussion}\label{sec:9}

In this paper, we began by reviewing tactile servo control and pose estimation, distinguishing image-based and pose-based control. We then formalised pose-based tactile servo control by blending classic concepts from visual servoing with tactile sensing using a PoseNet deep neural network for pose estimation. Our test robot comprised the BRL TacTip mounted on an industrial robot arm. These methods were validated with a quantitive assessment of tactile servoing performance on known regular objects (a sphere and disk) under changes in control parameters. We then demonstrated accurate servo control over many novel irregular objects, including a saddle, wave, several planar edges, a container lid, a bust of a human head and a squishy brain toy.  

\subsubsection*{Generality of the approach}
How broad a range of objects does our method of servo control apply to? Many different test shapes were accurately traced (\autoref{fig14}), using training data from just the edge and centre of a flat 3D-printed disk. The controller only struggled when the object became very curved. Although a soft sensor of this size (40\,mm dia. dome) cannot fully contact sharply concave surface features such as the base of the nose of the porcelain bust, this did not cause the servo control to fail (\href{https://youtu.be/N98nVqXvm88}{Supplementary Movie S7}).

The test surfaces varied in roughness, from smooth perspex and glazed porcelain to a textured plastic lid and 3D-printed stimuli. However, we did not notice any dependence of the servo control on roughness. In our view, this was because the effect of roughness was to change the shear on the sensor surface, and the PoseNets were trained to be insensitive to shear. For an initial test on a soft object, we tried tactile servoing over a `squishy brain'. The dynamics of the soft tactile sensor were found to be coupled to the object deformation, which is a rich topic for further investigation.

\subsubsection*{Pose-based and image-based servo control}
In our view, considering tactile servo control as pose-based or image-based will help the research field of robot touch progress in the future. The terminology PBTS and IBTS clarifies the relation with visual servoing, where the notion of PBVS and IBVS has been key for robot vision from the 1990s onwards. 

\textcolor{black}{How will this help robot touch progress? First, there is a wealth of knowledge from robot vision that can be transferred into robot touch, while appreciating the distinction between contact and non-contact modalities. Second, supervised deep learning with soft optical tactile sensors fits with PBTS, as pose information can be available during training data collection. However, IBTS control may be more appropriate in other situations; for example, image moments can be used to control planar tactile arrays without data-intensive training~\cite{li_control_2013,kappassov_touch_2020}. Also, unsupervised or semi-supervised deep learning may give other tactile image features that are suited for servo control. An appreciation of when to use PBTS or IBTS could give valuable insight into controlled touch for dexterous robots.}

\subsubsection*{Extensions to the approach}
One new direction would be to generalize the types of object feature beyond just either straight edges or surfaces. For example, tactile servo control has been applied to sliding along cables and rods~\cite{li_control_2013,kappassov_touch_2020} or switching between tactile servo controllers~\cite{kappassov_touch_2020}. In principle, a planner could set the object feature and reference during a task, such as to explore a smooth part of an object until reaching an edge then switch to tracing its boundary.


The present study considered position control because our industrial robot arm was limited to discrete pose updates at a latency of about 100\,ms.
Hence, our controller (Fig. 2) differs from other work on tactile servoing~\cite{li_control_2013,kappassov_touch_2020} that used a robot capable of velocity control. It would be straightforward to adapt our methods to velocity control to implement tasks such as tracking moving objects and smoothly servoing over surfaces and edges, using standard differential kinematics. In our view, a high frame-rate tactile sensor on a robot with millisecond update and latency could lead to servo control of a speed and precision far beyond that of human dexterity.

\begin{table*}[b]
	\renewcommand{\arraystretch}{1}
	\centering
	\begin{tabular}{@{}c|ccc@{}}	
		\textbf{object} & \textbf{\ samples\ } & \textbf{pose range} & \textbf{displacement range} \\
		\textbf{feature} & & \textbf{(labelled)} & \textbf{(unlabelled)} \\
		\hline
		3D surface & 10000 & $(0,0,[\text{-}5,\text{-}1];\pm15,\pm15,0)$ & $(\pm5,\pm5,0;\pm5,\pm5,\pm5)$ \rule{0pt}{2.5ex} \\
		3D edge & 10000 & $(\pm5,0,[\text{-}5,\text{-}1];\pm15,\pm15,\pm45)$ & $(\pm5,\pm5,0;\pm5,\pm5,\pm5)$ \rule{0pt}{2.5ex} \\
		\multicolumn{4}{c}{} 
	\end{tabular}
	\caption{Training/validation data parameters (units: mm, mm, mm; deg, deg, deg).}
	\label{tab:2}
	\begin{tabular}{@{}c|cccccc|cccc@{}}		
		\textbf{object} & \textbf{\# conv.} & \textbf{ \# dense } &\textbf{batch} & \textbf{\# conv.} & \textbf{\# dense} & \textbf{batch} & \textbf{activation} & \textbf{dropout} & \textbf{L1-reg.} & \textbf{L2-reg.} \\
		\textbf{feature} & \textbf{layers} & \textbf{layers } & \textbf{norm.} & \textbf{kernels} & \textbf{units} & \textbf{size} &\textbf{function}  & \textbf{coef.} & \textbf{coef.} & \textbf{coef.} \\
		\hline
		3D surface & 5 & 1 & None & 256 & 256 & 16 & ELU & 0.0047 & 0.0001 & 0.0159 \rule{0pt}{2.5ex} \\
		3D edge & 5 & 1 & None & 256 & 256 & 16 & ReLU & 0.0700 & 0.0001 & 0.0019 \rule{0pt}{2.5ex} \\
		\multicolumn{11}{c}{} 
	\end{tabular}
	\caption{PoseNet parameters. The first 6 parameters were fixed (using values from \cite{lepora_optimal_2020}); the final 4 parameters were optimized.}
	\label{tab:3}

	\begin{tabular}{@{}c|c|c|cc|c@{}}		
		\textbf{object} & \textbf{sensor pose} & \textbf{reference pose} & \textbf{proportional gain} & \textbf{integral gain} & \textbf{integral bounds} \\
		\textbf{feature} & ${}^FP_S$ & ${}^FP_R$ & ${\rm diag}(K_P)$ & ${\rm diag}(K_I)$ & $B$ \\
		\hline
		3D surface & $(0,0,z;\alpha,\beta,0)$ & $(0,1,\text{-}3;0,0,0)$ & $(1,1,\frac{1}{2};\frac{1}{2},\frac{1}{2},1)$ & $(0,0,\frac{3}{10};\frac{1}{10},\frac{1}{10},0)$ & $\pm(0,0,5;15,15,0)$ \rule{0pt}{2.5ex} \\
		3D edge & $(x,0,z;\alpha,\beta,\gamma)$  & $(0,1,\text{-}3;0,0,0)$ & $(\frac{1}{2},1,\frac{1}{2};\frac{1}{2},\frac{1}{2},\frac{1}{2})$ & $(\frac{3}{10},0,\frac{3}{10};\frac{1}{10},\frac{1}{10},\frac{1}{10})$ & $\pm(5,0,5;15,15,45)$ \rule{0pt}{2.5ex} \\ 
		\multicolumn{6}{c}{} 
	\end{tabular}
	\caption{Pose-based servo control parameters (units: mm, mm, mm; deg, deg, deg).}
	\label{tab:1}
\end{table*}

\section{Conclusion}

The pose-based tactile servo control described here enables a range of competencies that allow robots to interact physically with their environments. This includes any task that involves precise control of a soft tactile-sensorized part of a robot to have a desired pose relative to a surface, edge or other local object feature. This control can enable tactile probes to accurately explore and map objects or surfaces, which is applicable to metrology and materials testing; also, those probes can be used as non-prehensile manipulators to push or reposition objects~\cite{lloyd2021goal}. The control could also guide the precise use of fingertips on a robot hand to stabilize and explore held objects~\cite{psomopoulou2021robust}, which is a necessary component of in-hand manipulation. The ultimate aim of controlled soft touch is to provide a route towards an ease of dexterity that any manual task currently done by humans could be automated.

\section*{Supplementary material}

\noindent 
\href{https://youtu.be/PrtpVU-a1rk}{Movie S1 - Sphere surface} {{\tt\footnotesize {youtu.be/PrtpVU-a1rk}}\\
\href{https://youtu.be/UXIM8k3Pjqs}{Movie S2 - Disk edge} {{\tt\footnotesize {youtu.be/UXIM8k3Pjqs}}\\
\href{https://youtu.be/NldNjDKNS-I}{Movie S3 - Planar edged objects)}
 {{\tt\footnotesize youtu.be/NldNjDKNS-I}}\\
\href{https://youtu.be/nlFep15iZdE}{Movie S4 - Planar edged objects (2D)}
 {{\tt\footnotesize youtu.be/nlFep15iZdE}}\\
\href{https://youtu.be/7CfnEZGgBRc}{Movie S5 - Wave edge \& surface} {{\tt\footnotesize {youtu.be/7CfnEZGgBRc}}\\
\href{https://youtu.be/QC3Q6ArYoUE}{Movie S6 - Saddle edge and surface} {{\tt\footnotesize {youtu.be/QC3Q6ArYoUE}}\\
\href{https://youtu.be/N98nVqXvm88}{Movie S7 - Head surface} {{\tt\footnotesize {youtu.be/N98nVqXvm88}}\\
\href{https://youtu.be/x-xnLj_RgAk}{Movie S8 - Lid edges} {{\tt\footnotesize {youtu.be/x-xnLj\_RgAk}}\\
\href{https://youtu.be/BWrq5z0dLf4}{Movie S9 - Squishy brain surface} {{\tt\footnotesize {youtu.be/BWrq5z0dLf4}}\\
Linked as YouTube videos {{\tt\footnotesize {https://youtu.be/...}}

\appendix
Training data parameters, control parameters and PoseNet parameters used in this work are given in Tables~\ref{tab:2}-\ref{tab:1}.

\section*{Acknowledgment} 

This work was supported by an award from the Leverhulme Trust: `A biomimetic forebrain for robot touch' (RL-2016-39).

\bibliographystyle{unsrt}
\bibliography{references}

\end{document}